\documentclass[letterpaper, 10 pt, conference]{ieeeconf}  % Comment this line out if you need a4paper
\usepackage{graphicx}
\usepackage[linesnumbered,ruled,vlined]{algorithm2e}
\SetKw{KwBreak}{break}
\usepackage{algorithmicx}
\usepackage{amsmath}
\usepackage{mathtools}
\usepackage{booktabs}
\usepackage{array}
\usepackage{subcaption}
\usepackage{booktabs}
\usepackage{tabularx}
\usepackage{multirow}
\usepackage{threeparttable}
\usepackage{float}
\usepackage{svg}
\usepackage{amssymb}
\usepackage{amsfonts}
\usepackage{float}
\usepackage{comment}
\usepackage{url} 
\usepackage{balance}
\usepackage{cite}
\usepackage{multicol}
\usepackage[noend]{algpseudocode}
\usepackage[colorlinks=true, linkcolor=red, citecolor=red, urlcolor=red]{hyperref}

\IEEEoverridecommandlockouts                              % This command is only needed if 
\title{\LARGE \bf
MO-SeGMan: Rearrangement Planning Framework for Multi Objective Sequential and Guided Manipulation in Constrained Environments
}

\author{Cankut Bora Tuncer$^{1}$, Marc Toussaint$^{2}$, and Ozgur S. Oguz$^{1}$% <-this % stops a space
\thanks{$^{1}$LIRA Lab, Bilkent University, Turkey.}% <-this % stops a space
\thanks{$^{2}$LIS Lab, TU Berlin, Germany}% <-this % stops a space
\thanks{*This work was supported by TUBITAK under 2232 program with project number 121C148 (``LiRA").}% <-this % stops a space
\thanks{\textbf{Corresponding author:} Cankut Bora Tuncer, Department of Computer Engineering, Bilkent University, 06800 Bilkent, Ankara, Turkey. Email: {\tt\small bora.tuncer@bilkent.edu.tr}}%
}

\begin{document}

\maketitle
\thispagestyle{empty}
\pagestyle{empty}

%%%%%%%%%%%%%%%%%%%%%%%%%%%%%%%%%%%%%%%%%%%%%%%%%%%%%%%%%%%%%%%%%%%%%%%%%%%%%%%%
\begin{abstract}
In this work, we introduce MO-SeGMan, a Multi-Objective Sequential and Guided Manipulation planner for highly constrained rearrangement problems. MO-SeGMan generates object placement sequences that minimize both replanning per object and robot travel distance while preserving critical dependency structures with a lazy evaluation method. To address highly cluttered, non-monotone scenarios, we propose a Selective Guided Forward Search (SGFS) that efficiently relocates only critical obstacles and to feasible relocation points. Furthermore, we adopt a refinement method for adaptive subgoal selection to eliminate unnecessary pick-and-place actions, thereby improving overall solution quality. Extensive evaluations on nine benchmark rearrangement tasks demonstrate that MO-SeGMan generates feasible motion plans in all cases, consistently achieving faster solution times and superior solution quality compared to the baselines. These results highlight the robustness and scalability of the proposed framework for complex rearrangement planning problems. Supplementary videos and code are available at: \url{https://sites.google.com/view/mo-segman/}.
  
\end{abstract}

%%%%%%%%%%%%%%%%%%%%%%%%%%%%%%%%%%%%%%%%%%%%%%%%%%%%%%%%%%%%%%%%%%%%%%%%%%%%%%%%
\section{Introduction}

Rearrangement planning involves generating a feasible motion plan for a robot to move all goal objects to their designated locations.
A common strategy is to derive an object placement sequence from dependency relationships among objects and iteratively relocate obstructing objects to buffer locations when necessary.
Most prior work assumes that objects can be freely grasped and moved, an assumption that holds in typical tabletop rearrangement settings~\cite{hu2025mobile,gao2022fast,krontiris2016efficiently,gao2023minimizing,han2018complexity,han2017high,hu2024planning,gao2023orla,huang2024toward,ren2022rearrangement}.
However, this assumption often does not hold in cluttered environments, where objects severely restrict the robot’s reachable joint space, reducing the applicability of such methods.
Some recent studies incorporate joint-space constraints into rearrangement planning~\cite{ren2024multi,cheong2020relocate,levit2024solvingsequentialmanipulationpuzzles,tuncer2025segmansequentialguidedmanipulation,toussaint2024effort}, significantly improving feasibility in constrained settings.
Nonetheless, these approaches either rely on task or object specific heuristics that hinder generalization to diverse scenarios, or they fail to scale in highly cluttered environments with many redundant objects.

\begin{figure}[!t]
    \centering
    \resizebox{0.45\textwidth}{!}{\includegraphics{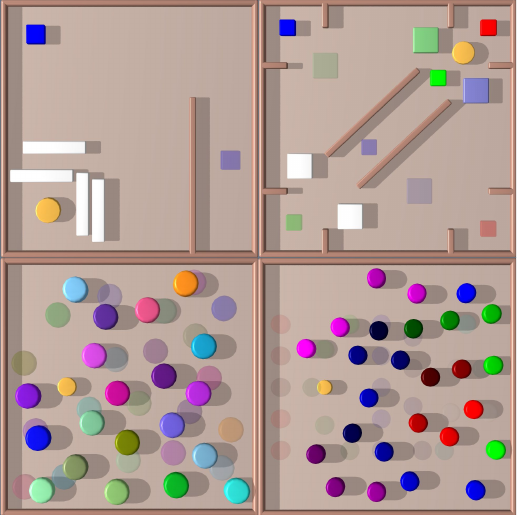}}
    \caption{Rearrangement planning cases of varying difficulty. 
    The objective is to generate a motion plan for the robot (yellow) to move the goal objects (colored) to their designated goal locations (silhouettes), while interacting with movable obstacles (white) and other goal objects.}
    
    \label{fig:intro}
\end{figure}

Several representative rearrangement scenarios are illustrated in Fig.~\ref{fig:intro}. In all scenarios, the robot's joint space is highly constrained by surrounding objects or movable obstacles. As a result, it may not be always possible to pick or place an object without first clearing obstructing objects. This makes object rearrangement particularly challenging, highlighting the importance of careful placement sequencing and robust manipulation planning.

In this paper, we introduce MO-SeGMan, the Multi-Objective Sequential and Guided Manipulation Planner. 
It addresses rearrangement problems where objects need to be arranged and manipulated in cluttered and highly constrained environments among movable obstacles. 
MO-SeGMan generates an object placement sequence for both monotonic and non-monotonic instances and iteratively manipulates the objects by decomposing the pick-and-place task into manageable subgoals while relocating obstructing obstacles. 

The main contributions of this work are:  
\begin{itemize}
    \item A sequence generation algorithm that jointly minimizes the replanning required per object and the robot’s overall travel distance using an iterative lazy evaluation method.  
    \item A scalable Selective Guided Forward Search (SGFS) for highly cluttered non-monotone scenarios, which relocates only critical obstacles while efficiently guiding the search toward configurations that both reduce obstructions and expand the robot’s feasible joint space.
    \item A subgoal refinement method that reduces unnecessary pick-and-place actions, improving overall solution quality.  
\end{itemize}

MO-SeGMan is evaluated on 14 rearrangement scenarios of increasing difficulty. 
Experimental results demonstrate its robustness and scalability, further validated through an ablation study. 
Moreover, MO-SeGMan achieves higher solution quality with reduced computation time compared to the baseline methods.

\section{Related Work}
Rearrangement planning is a widely studied problem, traditionally addressed through Task and Motion Planning (TAMP)\cite{srivastava2014combined} and Navigation Among Movable Objects (NAMO)\cite{schulman2014motion} frameworks, with recent works increasingly shifting focus toward highly constrained environments.
The main challenge arises from the combinatorial complexity of object placement orderings and the presence of movable obstacles, making the problem NP-Hard~\cite{wilfong1988motion, stilman2008planning}.
A common strategy for generating placement sequences is the use of dependency graphs~\cite{hu2025mobile,gao2022fast,krontiris2016efficiently,gao2023minimizing,han2018complexity,ren2024multi}. A sequence is generated from the topology of the dependency graph in particular, Han et al.~\cite{han2018complexity} further formulated the problem as a Traveling Salesman Problem (TSP), where optimized orderings are computed using a Euclidean distance heuristic. In non-monotone cases, some objects must be temporarily placed at external or internal relocation points, with the latter posing a greater challenge due to the limited space available for relocation.

Unlike TORO (Tabletop Rearrangement with Overhand Grasps)~\cite{hu2025mobile,gao2022fast,krontiris2016efficiently,gao2023minimizing,han2018complexity,han2017high,hu2024planning,gao2023orla,huang2024toward,ren2022rearrangement} approaches, where objects can be freely moved once picked, identifying feasible internal relocation points in cluttered, constrained environments is significantly harder due to the difficulty of finding motion plans for picking or placing objects in the presence of movable obstacles.
To address such rearrangement planning problems, where object relocation incorporates joint space limitations, several methods have been proposed.

In \cite{ren2024multi}, a Multi-Stage MCTS is integrated with buffer selection and motion planning, prioritizing relocations near shelf walls, whereas \cite{cheong2020relocate} filters feasible relocation points using a modified VFH+ combined with motion planning.
While robust in cluttered environments, both rely on object and case dependent heuristics, limiting the applicability of the methods.
\cite{bayraktar2023solving} introduced LA-RRT, which factors movable object state spaces and generates fragmented relocation plans that are later merged for efficiency.
\cite{levit2024solvingsequentialmanipulationpuzzles,toussaint2024effort} adopt Multi-Bound Tree Search (MBTS) for motion planning, with \cite{levit2024solvingsequentialmanipulationpuzzles} running a separate forward search obstacle relocations, yielding a more generalizable framework.
However, these methods struggle to scale in highly cluttered settings with many redundant objects, as they expand the search across all possible relocations. Unlike these approaches, SeGMan~\cite{tuncer2025segmansequentialguidedmanipulation} reduces the search space by identifying critical object sets for relocation and exploring them via best-first search over multiple trees.
Relocation points are selected from free-space regions independent of object types, and obstacles are iteratively relocated until feasible task trajectories are found.
SeGMan also employs a hybrid motion planning method that combines sampling- and optimization-based methods, adaptively refining object subgoals to handle environmental constraints.
Although SeGMan scales better than\cite{toussaint2024effort} and\cite{levit2024solvingsequentialmanipulationpuzzles}, its performance degrades when the number of movable objects exceeds $m > 10$, where it still struggles to find feasible motion plans.

To address the generalizability and scalability issues of the recent works which does rearrangement planning in constrained environments, MO-SeGMan is presented.

\section{Problem Formulation}
We assume a configuration space $\mathcal{X} \subset \mathbb{R}^n \times SE(3)^m$ of an $n$-DoF robot, $m$ movable objects $\mathcal{O} = \{o_1,\dots,o_m\}$, goal objects $O_g \subseteq \mathcal{O}$ and their goal locations $\mathcal{G \in \mathbb{R}}$. 
The object placement motion plan for each goal object is defined as 
\[ 
\tau_i = \big(o_i, \tau^{\text{pick}}_i, \tau^{\text{place}}_i\big)
\]
where $\tau^{\text{pick}}_i : [0,1] \to \mathcal{X}$ and $\tau^{\text{place}}_i : [0,1] \to \mathcal{X}$ denote the pick and place trajectories, respectively. 
To ensure continuity, the following conditions are imposed:
\[
\forall i: \quad \tau^{\text{pick}}_i(1) = \tau^{\text{place}}_i(0), 
\quad 
\tau^{\text{pick}}_{i+1}(0) = \tau^{\text{place}}_i(1) 
\]
so that consecutive pick-and-place actions are connected smoothly.

The $\tau^{\text{pick}}_i$ and $\tau^{\text{place}}_i$ are generated using the k-order Markov Motion Optimizer (KOMO) \cite{14-toussaint-KOMO} and the bi-directional Rapidly-exploring Random Trees (RRTs)~\cite{kuffner2000rrt}.    
KOMO formulates the motion plan as a nonlinear constrained optimization problem over trajectories in configuration space. 
Let $x_t \in \mathbb{R}^n$ be the configuration at time $t$, and $x_{0:T} = (x_0, \dots, x_T)$ denote a trajectory of horizon $T$. 
KOMO solves a $k$-order constrained non-linear problem of the form:
\begin{align}
\min_{x_{0:T}} \quad & \sum_{t=0}^{T} f_t(x_{t-k:t})^\top f_t(x_{t-k:t}) \label{eq:komo} \\
\text{s.t.} \quad & g_t(x_{t-k:t}) \leq 0, \quad h_t(x_{t-k:t}) = 0, \quad \forall t \nonumber
\end{align}
where $f_t$, $g_t$, and $h_t$ are differentiable cost, inequality, and equality constraint functions, respectively, defined over tuples of $k+1$ consecutive states $x_{t-k:t}$.
The constraints \textit{touch}, \textit{stable}, and \textit{positionDiff} are grounded as geometric constraints \cite{toussaint2015logic}. 
The \textit{touch} constraint enforces contact between the robot end-effector and the object’s grasp point, modeled as an equality constraint on their Euclidean distance. 
The \textit{stable} constraint ensures that an object remains fixed at a relative pose across timesteps, preventing unintended motion once placed. 
The \textit{positionDiff} constraint bounds the relative displacement between two bodies, and represented as an inequality. 
KOMO first finds a feasible pick configuration $x_i^{\text{pick}}$, and Bi\mbox{-}RRT generates a $\tau_{i-1}^{\text{place}}(1)$ from $x_i^{\text{pick}}$ to obtain $\tau_i^{\text{pick}}$. To generate $\tau_i^{\text{place}}$, intermediate object placement subgoals are determined along the object trajectory $\mu_i^{\text{place}}\subset SE(3)$ with Bi\mbox{-}RRT. KOMO then optimizes a motion plan for the robot that iteratively places the object at these subgoals, yielding $\tau_i^{\text{place}}$.

The aim is to generate a sequence of manipulations $\{\tau_i\}_{i=1}^K$ while minimizing $K$, 
where $K$ denotes the total number of object interactions. 
An object interaction is defined as an iterative sequence of pick-and-place actions, counted both when a goal object is moved to its target location and when an obstacle is relocated. 
To minimize $K$, the object placement sequence $\mathcal{S}$ is followed. 
For \textit{monotone} cases, $\mathcal{S}$ is a permutation of the goal objects $O_g \subseteq \mathcal{O}$, where each object is placed with exactly one interaction. 
In \textit{non-monotone} cases, $\mathcal{S}$ is not a strict permutation of $O_g$ but may contain repeated entries of the same goal object, reflecting temporary relocations required to resolve dependency cycles. The algorithm terminates when all goal objects reach their designated targets, i.e., 
$\forall o_g \in O_g : x^{\text{final}}_{o_g} = g_{o_g}$.

\setlength{\textfloatsep}{5pt plus 1.0pt minus 2.0pt}
\begin{algorithm}[!t]
\caption{MO-SeGMan}
\label{alg:mosegman}
\SetInd{0.5em}{0.5em}
\footnotesize
\DontPrintSemicolon
\SetAlgoNoLine
\SetAlgoNoEnd
\KwIn{Initial configuration $x_0$, goal set $O_g$}
\KwOut{Motion plan $\tau$}
\textbf{Initialize:} $\tau \gets \varnothing$, $skip \gets 0$, $skip_{\max} \gets \lfloor |O_g|/4 \rfloor$, $iter \gets 0$, $iter_{\max} \gets |O_g|+1$ \;
$\mathcal{S} \gets \texttt{genObjPlaceSeq}(x_0, O_g)$ \Comment{Sec.\ref{sec:objplacseqgen}}\;
$O_{placed} \gets \texttt{verifObjPlacement}(x_0)$ \;
$x\gets x_0$ \;
\While{$|O_{placed}| \neq |O_g| \land iter < iter_{\max}$}{
  $iter \gets iter + 1$ \;
  \For{$o_i \in \mathcal{S}$}{
    \If{$o_i$ not placed}{
      $\tau_i, x' \gets \texttt{genMotionPlan}(o_i, x, O_{placed})$\Comment{Sec.\ref{sec:manipplan}}\;
      \If{$\tau_i$ feasible}{
        $skip \gets 0$ \;
        $x \gets x'$ \;
        $\tau \gets \tau \cup \{\tau_i\}$ \;
        $O_{placed} \gets \texttt{verifObjPlacement}(x)$ \;
        $O_{notPlaced} \gets O_g \setminus O_{placed}$ \;
        \If{$\texttt{anyReloc}(x)$}{
          $\mathcal{S} \gets \texttt{genObjPlaceSeq}(x, O_{notPlaced})$ \;
          \KwBreak
        }
      }
      \Else{
        $skip \gets skip + 1$ \;
        \If{$skip > skip_{\max}$}{
          $skip \gets 0$ \;
          $O_{notPlaced} \gets O_g \setminus O_{placed}$ \;
          $\mathcal{S} \gets \texttt{genObjPlaceSeq}(x, O_{notPlaced})$ \;
          \KwBreak
        }
      }
    }
  }
}
\Return{$\tau$}
\end{algorithm}

\section{Multi-Objective Sequential and Guided Manipulation Planning}
The main algorithm for MO-SeGMan is presented in Alg.~\ref{alg:mosegman}. 
Given an initial configuration $x_0 \in\mathcal{X}$, the MO-SeGMan attempts to place the goal objects at goal locations. To minimize the planning required per object and the robots overall travel distance, the algorithm starts by generating an optimized object placement sequence $\mathcal{S}$ which is described in Sec.~\ref{sec:objplacseqgen}. 
The algorithm proceeds until all goal objects are placed or the iteration limit is reached. 
Following the order in $\mathcal{S}$, a motion plan $\tau_i$ consisting of sequential pick-and-place actions is generated (Sec.~\ref{sec:manipplan}), relocating movable obstacles when necessary with Selective Guided Forward Search (SGFS) (Sec.~\ref{sec:sgfs}). This iterative placement continues until all the goal objects in $\mathcal{S}$ is placed.

Since manipulation planning dominates the runtime, updating $\mathcal{S}$ when needed is a practical trade-off: re-optimizing $\mathcal{S}$ reduces costly future replanning by adapting to the evolving configuration. For example, when a goal object is temporarily relocated, it may alter the dependency relationships in $G'$, requiring $\mathcal{S}$ to be regenerated for the remaining objects. Moreover, if a feasible motion plan $\tau_i$ cannot be generated for an object in $\mathcal{S}$, its placement is skipped and retried later. When more than $skip_{\max}$ consecutive objects are skipped, $\mathcal{S}$ is regenerated to reflect changes in $x$, such as updated robot or object positions.

\begin{figure}[t]
\centering
\resizebox{0.3\textwidth}{!}{\includegraphics{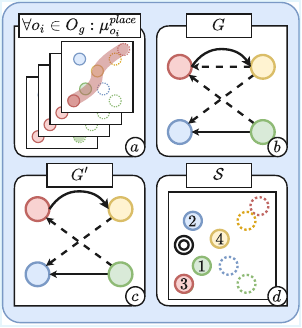}}
\caption{Steps in object placement sequence generation: (a) collision checking along the object placement trajectories $\mu_{o_i}^{\text{place}}$, (b) the dependency graph $G$ (dotted edges: weak dependencies, solid edges: strong dependencies), (c) acyclic dependency graph $G'$, and (d) optimized object placement sequence $\mathcal{S}$.}
\label{fig:SeqOpt}
\end{figure}

\subsection{Object Placement Sequence Generation}
\label{sec:objplacseqgen}
In this phase, object placement sequence $\mathcal{S}$ is determined for the $O_g$. 
While determining the order, the aim is to jointly minimize the required motion plan per object as well as the distance covered by the robot. The critical components of the sequence generation are cached, so that during regeneration of $\mathcal{S}$ they do not need to be recomputed from scratch.
It follows a two step procedure, dependency graph generation, and sequence optimization.

\subsubsection{\textbf{Dependency Graph Generation}}
Object dependencies are represented by a directed graph $G = (V,E)$, where vertices $V$ denote goal objects and each edge $e_{i,j} \in E$ encodes that $o_i$ must be placed before $o_j$. 
The graph $G$ is constructed by identifying collisions along the object placement trajectories $\mu_{o_i}^{\text{place}}$ for all $o_i \in O_g$, generated in an auxiliary configuration where all other movable objects $\mathcal{O} \setminus \{o_i\}$ are removed (Fig. \ref{fig:SeqOpt}a).  

Type~1 collisions occur when $\mu_{o_i}^{\text{place}}$ intersects the initial positions of other objects, requiring those objects to be relocated in advance. 
Type~2 collisions occur when $\mu_{o_i}^{\text{place}}$ intersects the goal locations of other objects, enforcing that the corresponding objects must be placed after $o_i$. 
Type~1 and Type~2 collisions are encoded in $G$ as \emph{weak} and \emph{strong} dependencies, respectively (Fig. \ref{fig:SeqOpt}b). 
Violating a strong dependency requires relocating an object and subsequently replacing it back to its goal, whereas violating a weak dependency only requires relocation. 
Hence, the motion planning cost of a strong dependency violation is double to a weak one.

The dependency graph often contains cycles (non-monotonic), particularly in cluttered environments. 
Such cycles must be resolved; otherwise, it is impossible to derive $\mathcal{S}$ that respects the precedence relationship between objects. 
For that purpose, $G$ is transformed into a Directed Acyclic Graph (DAG) $G'$ with a Depth-First Search (DFS) based method (Fig.\ref{fig:SeqOpt}c).
When a backtrack is detected, a cycle is identified and the corresponding edges are appended to a list with a unique cycle ID. 
After all cycles are enumerated, the edges participating in cycles are ranked by their frequency of occurrence and iteratively removed in decreasing order until no cycles remain in the graph. 
If multiple edges share the same cycle count, the edge with the weaker dependency is removed. 
If both are weak or strong edges, ties are broken by removing the edge with the smallest out-degree to in-degree difference. 

Compared to existing cycle removal techniques such as \cite{even1998approximating}, \cite{ariffin2022transformation}, or simple greedy elimination, the proposed method has higher computational complexity, growing exponentially with the number of cycles due to explicit enumeration. 
However, in practical scenarios with $m \leq 50$ objects, the cycle removal step remains practical, as overall runtime is dominated by manipulation planning rather than sequence optimization. 
The proposed approach is preferred over other cycle removal methods as it preserves critical dependency structures as much as possible, thereby minimizing the motion planning effort required per object.

\begin{figure}[t]
  \centering
   \resizebox{0.45\textwidth}{!}{\includegraphics{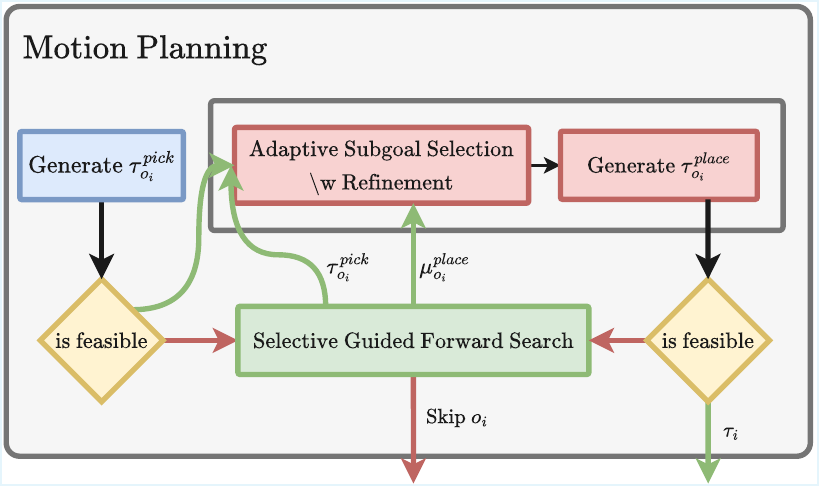}}
    \caption{Flowchart of the motion planning process. The pick and placement plan are decomposed due to the complex nature of the problem. If a motion plan cannot be generated, the movable obstacles are relocated with SGFS.}
  \label{fig:motplan}
\end{figure}

\subsubsection{\textbf{Sequence Optimization}}
From $G'$, precedence constraints for each object are derived directly from the graph topology.
Although these constraints restrict the set of valid placement orderings, they rarely yield a unique sequence.
To select an ordering that minimizes the robot’s travel distance, the problem is formulated as an Asymmetric Traveling Salesman Problem (ATSP) and solved via Integer Programming~(IP) (Fig.~\ref{fig:SeqOpt}d).
%For simplicity, travel during object placement is omitted, as once an object is picked the robot’s only destination is its designated goal location.
The initial edge costs are defined as the Euclidean distances 
\[
c_{ij} = \| g_{o_{i}} - o_{j} \|_2, \quad \forall\, i,j \in \{1,\dots,|O_g|\}, \; i \neq j
\]
which provides a loose lower bound.
A tighter estimate is obtained using RRT distances, which better approximate the true travel distance but are computationally more expensive. 
To balance accuracy and efficiency, a lazy evaluation strategy is adopted: an initial sequence $\mathcal{S}$ is computed with Euclidean costs, then refined by progressively updating edges with RRT distances. During re-optimization, it is warm started with the current best object sequence. 
This process repeats until $\mathcal{S}$ converges or the loop limit is reached, ensuring that the final sequence is optimal with respect to the cost matrix and precedence constraints.

\subsection{Motion Planning}
\label{sec:manipplan}
The motion plan generation process is illustrated in Fig.~\ref{fig:motplan}. Each motion plan $\tau_i$ consists of consecutive pick-and-place actions. 
For each object $o_i$, a feasible pick configuration $x_{o_i}^{\text{pick}}$ is checked (Eq.~\ref{eq:komo}), after which a motion plan $\tau_{o_i}^{\text{pick}}$ is generated using Bi-RRT such that $\tau_{o_i}^{\text{pick}}(1) = x_{o_i}^{\text{pick}}$, similar to~\cite{hartmann2022long}. 
If $\tau_{o_i}^{\text{pick}}$ is feasible, subgoals are then adaptively selected along the object trajectory $\mu_{o_i}^{\text{place}}$~\cite{tuncer2025segmansequentialguidedmanipulation}. 
In constrained regions, denser subgoals are chosen to allow multiple pick-and-place actions, while in free space they are placed more sparsely. 

Although adaptive subgoal selection\cite{tuncer2025segmansequentialguidedmanipulation} adjusts to the environment, a key drawback is that after clearing a narrow passage, the algorithm continues with unnecessarily small steps before increasing the step size. This leads to redundant intermediate pick-and-place sequences. 
To address this, we introduce a refinement step. 

Let $cp(\tau_{o_g}^{\text{place}}(k)) \subset SE(3)$ denote the contact point between $o_g$ and the robot at step $k$. 
Two consecutive steps $k$ and $k+1$ are merged if $\|cp(\tau_{o_g}^{\text{place}}(k)) - cp(\tau_{o_g}^{\text{place}}(k+1))\|_2 < \epsilon$ and the merge is feasible. 
This refinement reduces unnecessary pick-and-place operations, especially during transitions from narrow passages to free space.

\begin{figure}[t]
   \centering
    \resizebox{0.47\textwidth}{!}{\includegraphics{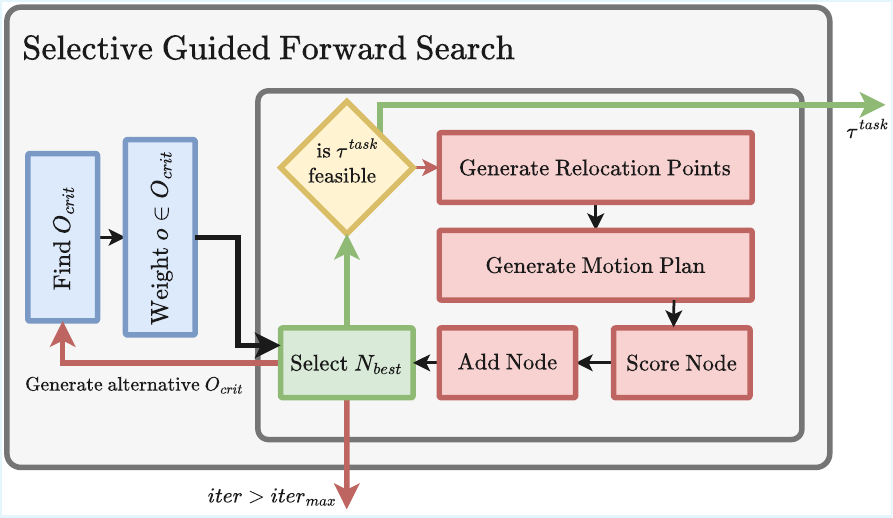}}
    \caption{The Selective Guided Forward Search (SGFS) explores configurations for the relocation of critical objects to obtain a feasible task trajectory.}
   \label{fig:sgfs}
\end{figure}

\subsection{Selective Guided Forward Search (SGFS)}
\label{sec:sgfs}
When a feasible $\tau_{o_i}^{\text{pick}}$ or $\mu_{o_i}^{\text{place}}$ - jointly referred to as the task trajectory $\mu_{o_i}^{\text{task}}$ - cannot be realized at configuration $x$, it is typically due to obstructing movable obstacles. 
To address this, we propose a scalable and robust forward search method, SGFS, which identifies critical obstacles and relocates them to feasible locations (Fig.~\ref{fig:sgfs})

The algorithm begins by identifying a subset of critical obstacles $O_{\text{crit}} \subset \mathcal{O}$ (Sec.~\ref{sec:critobjsel}), since searching over all movable objects would severely limit efficiency and scalability. 
Each critical object is assigned a relocation weight $w_r$, which determines the number of relocation points proposed for that object (Sec.~\ref{sec:objweight}). 
A list $L$ is then initialized with a root node $N_0 = (O_{\text{crit}}, x)$, where each node $N_i \in L$ is represented as $N_i = (O_i, x_i)$, with $O_i$ denoting the objects selected for relocation and $x_i$ the corresponding configuration. 

At each iteration, the highest-scoring node $N_{\text{best}} \in L$ is selected for expansion. 
For each reachable object in $O_{\text{best}}$, relocation points are generated (Sec.~\ref{sec:relocgen}), and motion plans are computed for placing the object at these points (Sec.~\ref{sec:manipplan}). 
Each feasible relocation yields a new node with an updated configuration, which is then scored (Sec.~\ref{sec:nodescore}). 
Finally, the top-scoring nodes are added to $L$.
 
The search terminates when a feasible task trajectory can be generated from $x_{\text{best}}$. 
If no task trajectory can be found within the iteration limit, another SGFS attempt is made with an alternative critical object set. 
If this also fails, MO-SeGMan skips the placement of $o_i$ and later reattempts to place it.  

\begin{figure}[t]
   \centering
   \resizebox{0.35\textwidth}{!}
   {\includegraphics{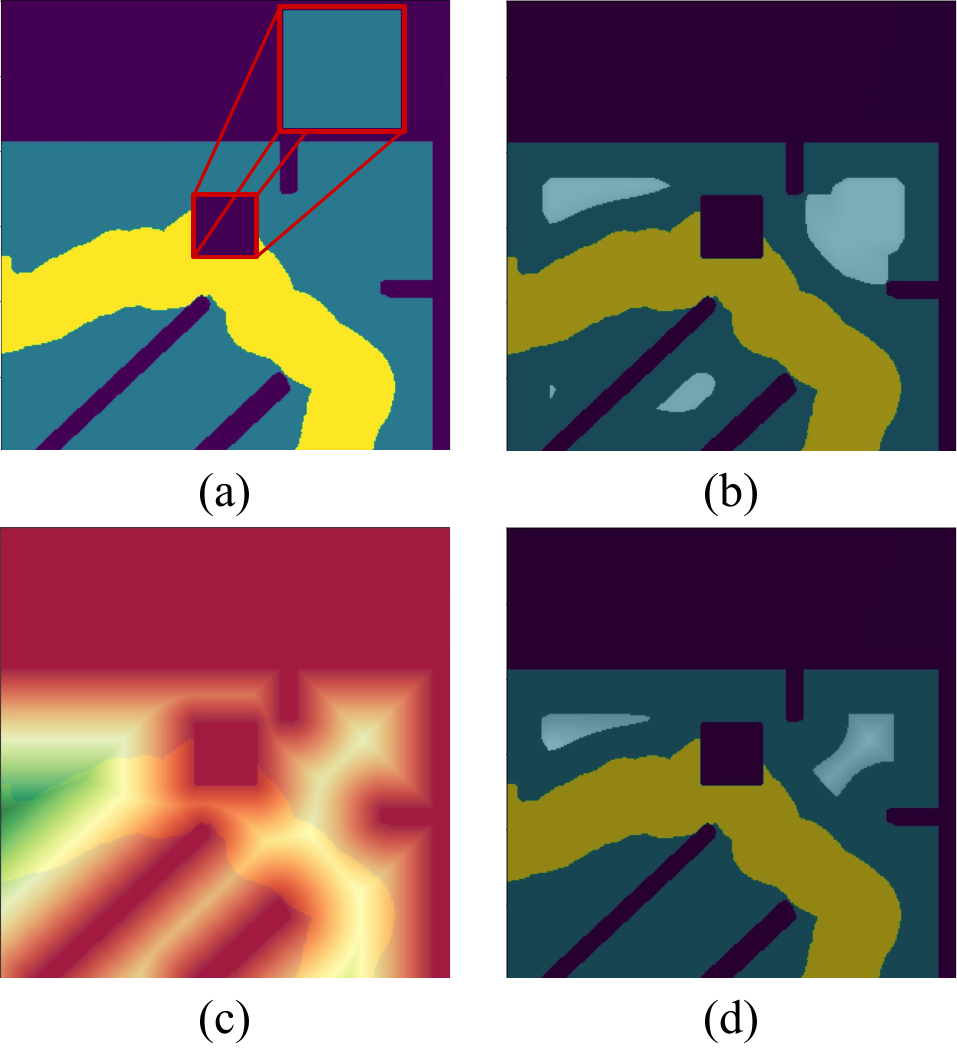}}
\caption{Relocation-point generation: (a) Local Occupancy Matrix (LOM) with object mask (red box), (b) convolution result (light areas), (c) Euclidean Distance Transform (green = higher clearance), (d) candidate relocation points (light areas).}
\label{fig:relocgen}
\end{figure}

\subsubsection{\textbf{Critical Object Selection}}
\label{sec:critobjsel}
The critical object set $O_{\text{crit}} \subset \mathcal{O}$ is defined as the minimal subset of objects that must be relocated to obtain a feasible task trajectory.
To obtain $O_{\text{crit}}$, we first identify the colliding objects $O_{\text{col}}$ along the task trajectory. 
Since $O_{\text{col}}$ does not necessarily yield the minimal subset - because some objects can be circumvented - an iterative elimination procedure is applied to filter out redundant obstacles.

\begin{figure}[t]
   \centering
   \resizebox{0.4\textwidth}{!}{\includegraphics{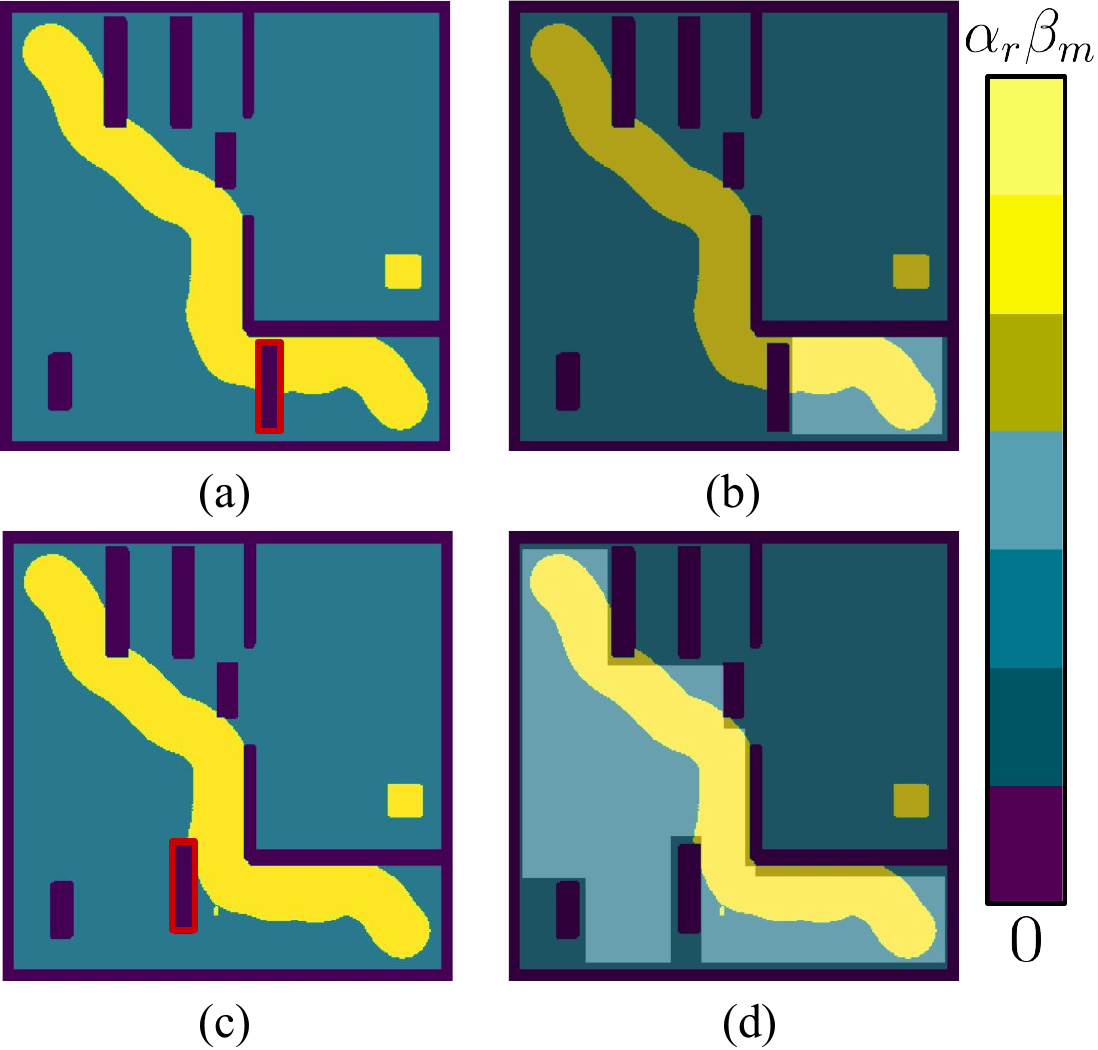}}
    \caption{Scene scoring process. (a) Global Occupancy Matrix (GOM) from the initial scene image, (b) scene score combined with the reachability matrix (lighter areas indicate higher reachability), (c–d) as obstacle (red) is relocated, the scene score increases.}
   \label{fig:scenescore}
\end{figure}

For each candidate subset $O_{\text{cand}} \subseteq \mathcal{P}(O_{\text{col}})$, where $\mathcal{P}(\cdot)$ denotes the power set, the objects in $O_{\text{cand}}$ are temporarily removed from $x$, and the feasibility of the task trajectory is evaluated. 
If the trajectory becomes feasible, the objects in $O_{\text{cand}}$ are classified as critical and designated as $O_{\text{crit}}$.

{
\begin{figure*}[!th]
  \centering
  \resizebox{0.97\textwidth}{!}{\includegraphics{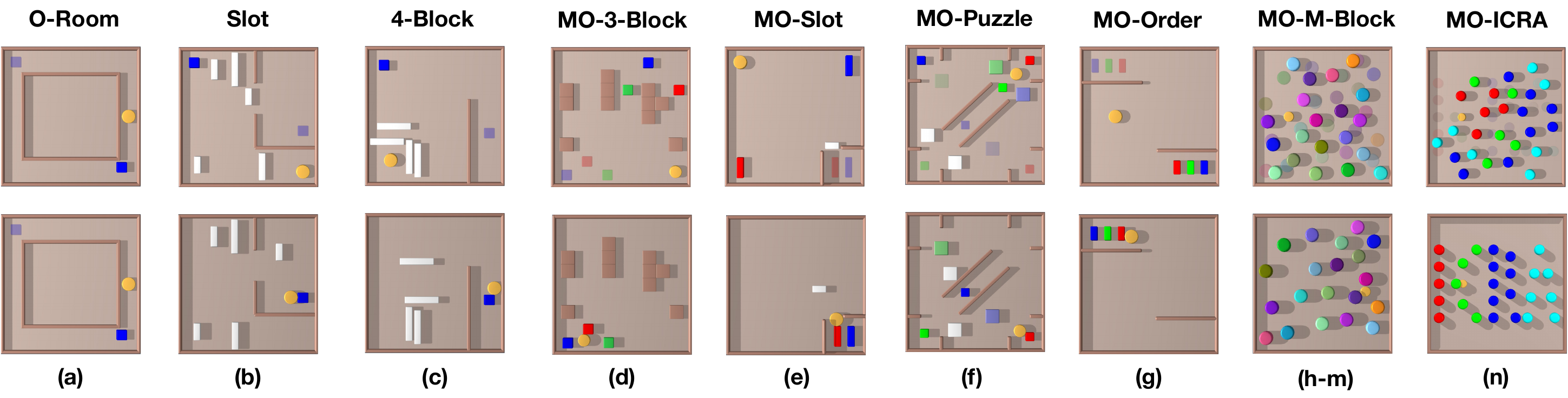}}
  \caption{MO-SeGMan evaluated on 14 (8 + 6) rearrangement tasks of varying difficulty. 
The first row shows the initial configurations, and the second row shows the desired goal configurations with all goal objects placed. The MO-M-Block is tested for different randomly placed object numbers ($M \in {2, 4, 8, 12, 16, 20}$).}
  \label{fig:task}
\end{figure*}
}

\subsubsection{\textbf{Node Scoring}}
\label{sec:nodescore}
To balance exploitation of promising nodes with sufficient exploration, the node scoring function combines a scene score with an exploration factor. 
The scene score is derived from the image of the configuration $x$, denoted $I(x)_{} \in \mathbb{R}^{h \times w}$. 

The Global Occupancy Matrix $M(I) = [m_{ij}]$ encodes the spatial layout of $x$: obstacles and objects are assigned $0$, free space $\alpha_m$, and robot or object placements along the task trajectory $\beta_m$, with $\beta_m > \alpha_m$ (Fig.~\ref{fig:scenescore}). 
Thus, larger values of $\sum m_{ij}$ correspond to configurations where the task trajectory is less obstructed.  

To incorporate joint-space constraints, the reachability matrix $R(I) = [r_{ij}]$ is computed via wavelet-style propagation from the robot’s current configuration. 
Elements are weighted $\alpha_r$ for reachable points and $\beta_r$ for unreachable points, with $\alpha_r > \beta_r$.  

The scene score is then defined as
\[
s_{\text{scene}}(x) = \sum_{i=1}^{h} \sum_{j=1}^{w} m_{ij}\, r_{ij},
\]
favoring configurations that both clear the task trajectory and expand the robot’s effective workspace.  

The exploration factor and overall node score are defined as
\[
\xi_{\text{exp}}(N) = c_0 \sqrt{\text{visit}(N)},
\quad
s_{\text{node}}(x) = s_{\text{scene}}(x) + \xi_{\text{exp}}(N)
\]
where $c_0$ is a constant and $\text{visit}(N)$ is the visitation count of node $N$.  

Hence, node selection prioritizes configurations that free the task trajectory and enlarge the robot’s workspace, while balancing exploration of less-visited nodes.

\subsubsection{\textbf{Object Weighting}}
\label{sec:objweight}
To dynamically control how many relocation points are generated for each $o \in O_{\text{crit}}$, objects are assigned a normalized relocation weight $w_r \in [0,1]$. 
These weights determine the allocation of search effort across different obstacles.  

Weights are computed by estimating the potential improvement in the scene score $s_{\text{scene}}(x)$ after relocating the object. 
Specifically, for each object it is assumed to be placed at a free-space location away from the task trajectory, and the change between the initial and updated scene score is measured. 
The difference is the best-case contribution of the object to the scene score after being relocated. 
Finally, these contributions are normalized across all objects in $O_{\text{crit}}$ to obtain the weights $w_r$.  
During the best node selection, the weight of the recently relocated object in $N_{prev}$ is reduced proportionally to the contribution to the configuration score, shifting the search focus toward objects that continue to obstruct the task trajectory. 

\subsubsection{\textbf{Relocation Point Generation}}
\label{sec:relocgen}
Relocation points are selected from feasible free-space locations near the object that are sufficiently distant from clutter. 
This ensures that objects are placed away from both the current task trajectory and other goal objects, preventing future obstructions (Fig.~\ref{fig:relocgen}).

Similar to scene scoring, the configuration image $I(x)$ is used. 
After encoding the spatial layout of $x$ with the Global Occupancy Matrix $M(I)$, a bounding window around the object is extracted to obtain the Local Occupancy Matrix (LOM), $L(I)$. 
The Euclidean Distance Transform (EDT) is then applied to $L(I)$, producing a clearance map $C(L(I))$ that measures the distance of each cell to the nearest obstacle. 
Feasible relocation points are finally obtained by convolving the free-space regions of $L(I)$ with the object mask and discarding indices with low matching scores or low clearance.  

\subsubsection{\textbf{Expanding $O_{crit}$}}
\label{sec:expand}
In some cases, relocation of objects in $N_{best}$ may fail because those objects are themselves obstructed by other obstacles.
Such obstacles are not part of the initial $O_{\text{crit}}$ and are instead incorporated dynamically.
When no improvement is observed in the node score, the object with the highest accumulated collision count during relocation attempts is added to $O_{\text{crit}}$ with a $w_r$ assigned proportional to its size.

\begin{figure*}[!th]
  \centering
  
  \begin{minipage}{0.173\textwidth}
    \centering
    \includegraphics[width=\textwidth]{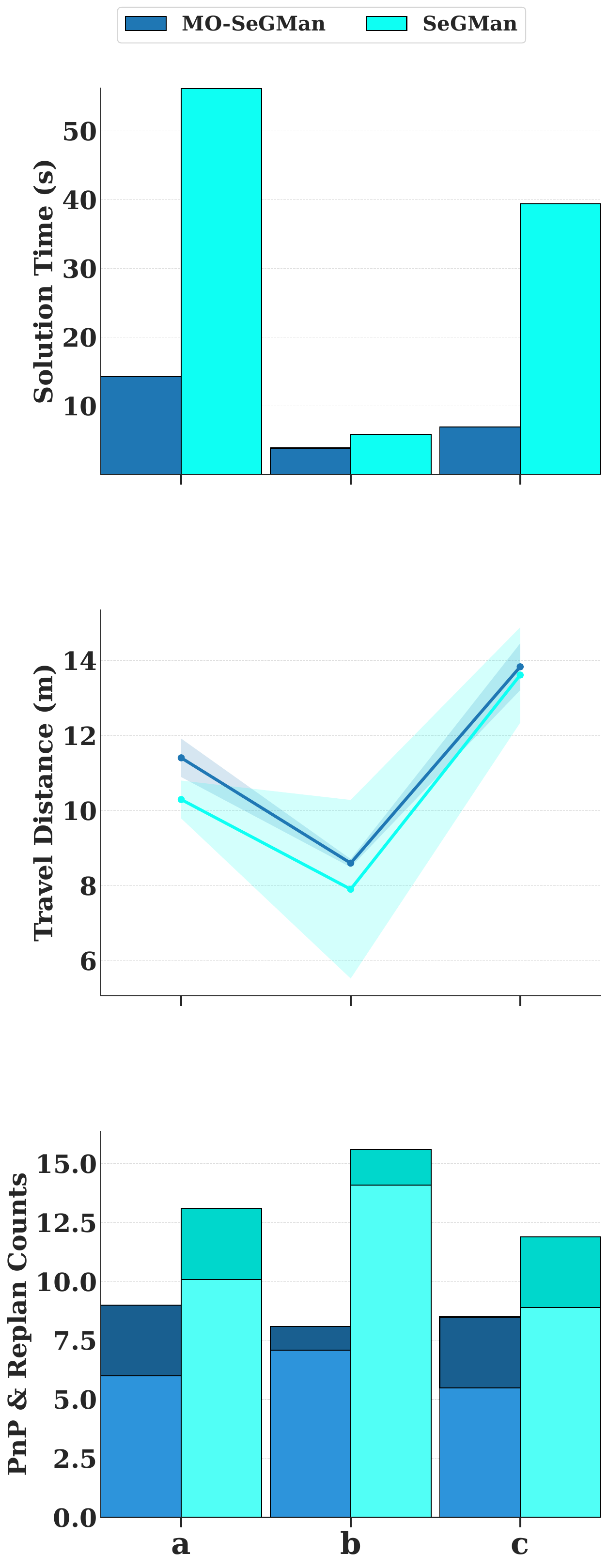}
  \end{minipage}%
  \begin{minipage}{0.17\textwidth}
    \centering
    \includegraphics[width=\textwidth]{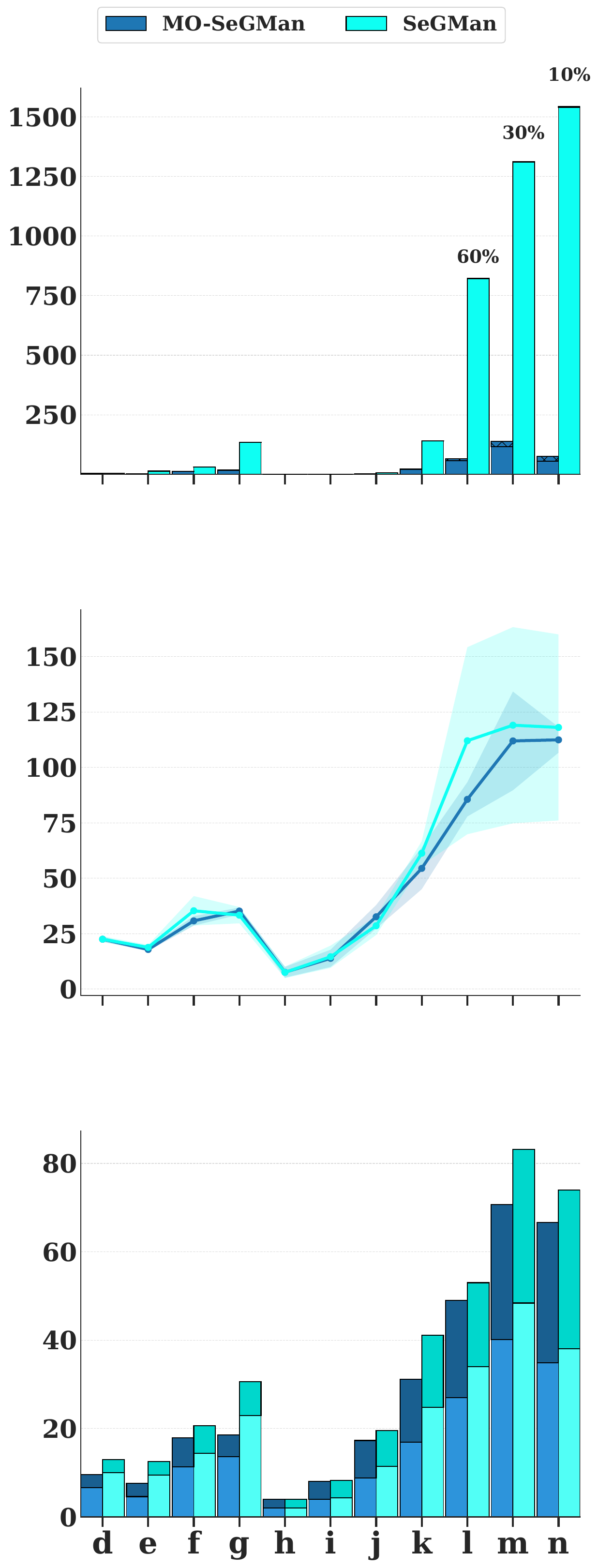}
  \end{minipage}%
  \begin{minipage}{0.16\textwidth}
    \centering
    \includegraphics[width=\textwidth]{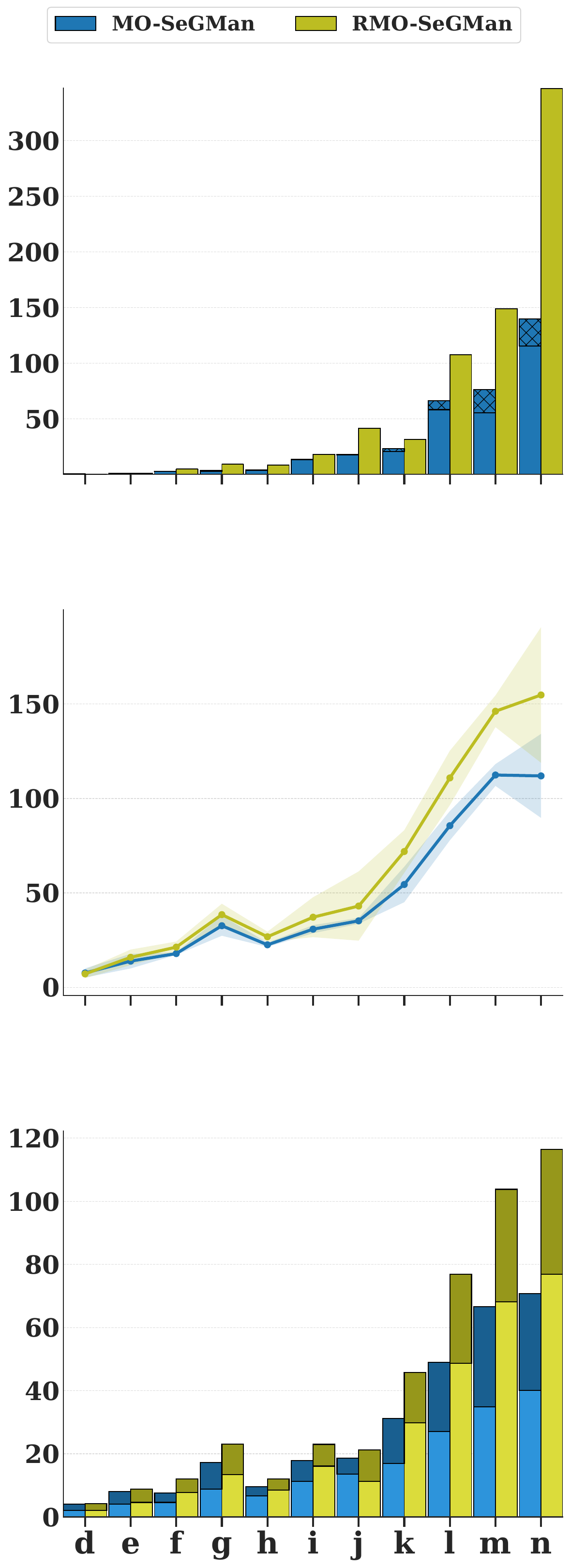}
  \end{minipage}%
  \begin{minipage}{0.16\textwidth}
    \centering
    \includegraphics[width=\textwidth]{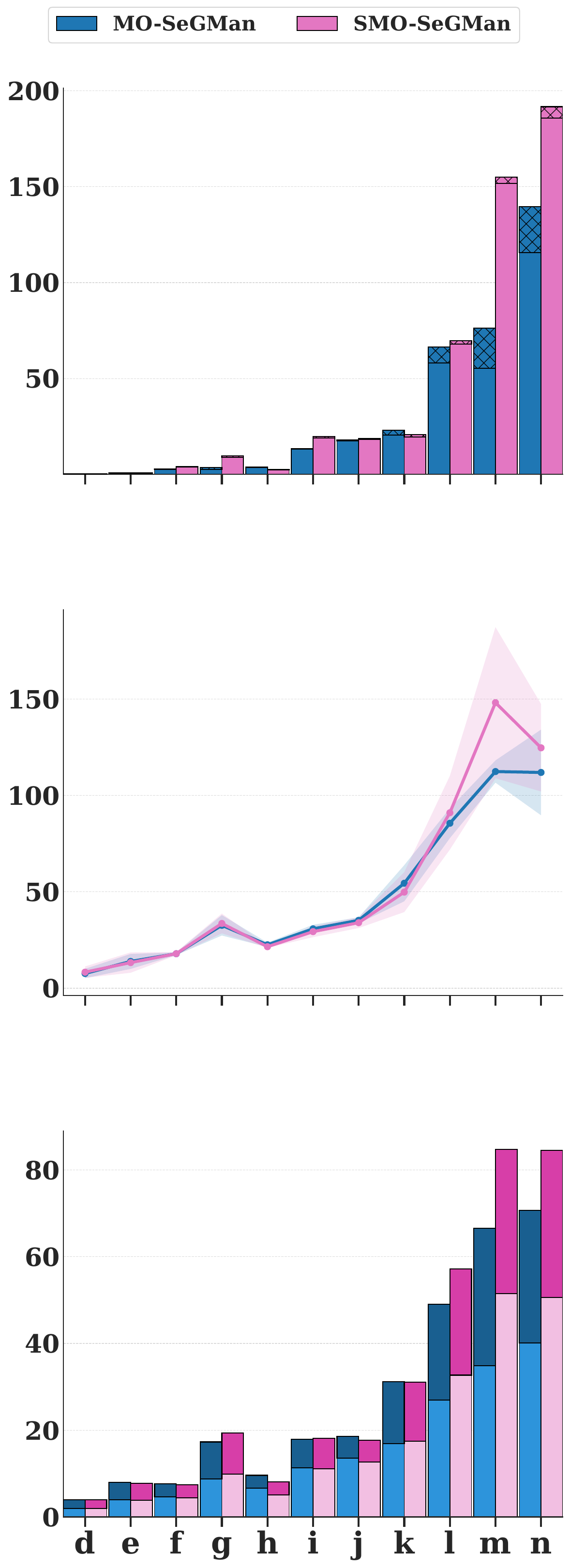}
  \end{minipage}%
  \begin{minipage}{0.16\textwidth}
    \centering
    \includegraphics[width=\textwidth]{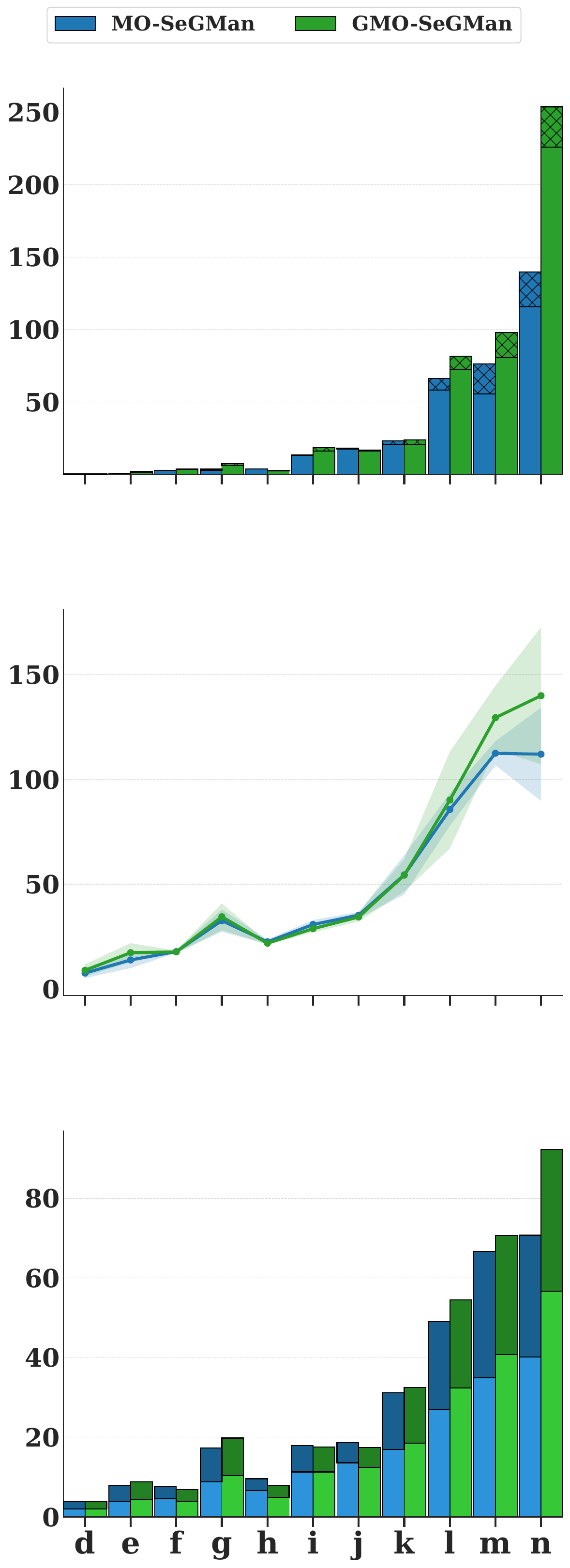}
  \end{minipage}%
  \begin{minipage}{0.16\textwidth}
    \centering
    \includegraphics[width=\textwidth]{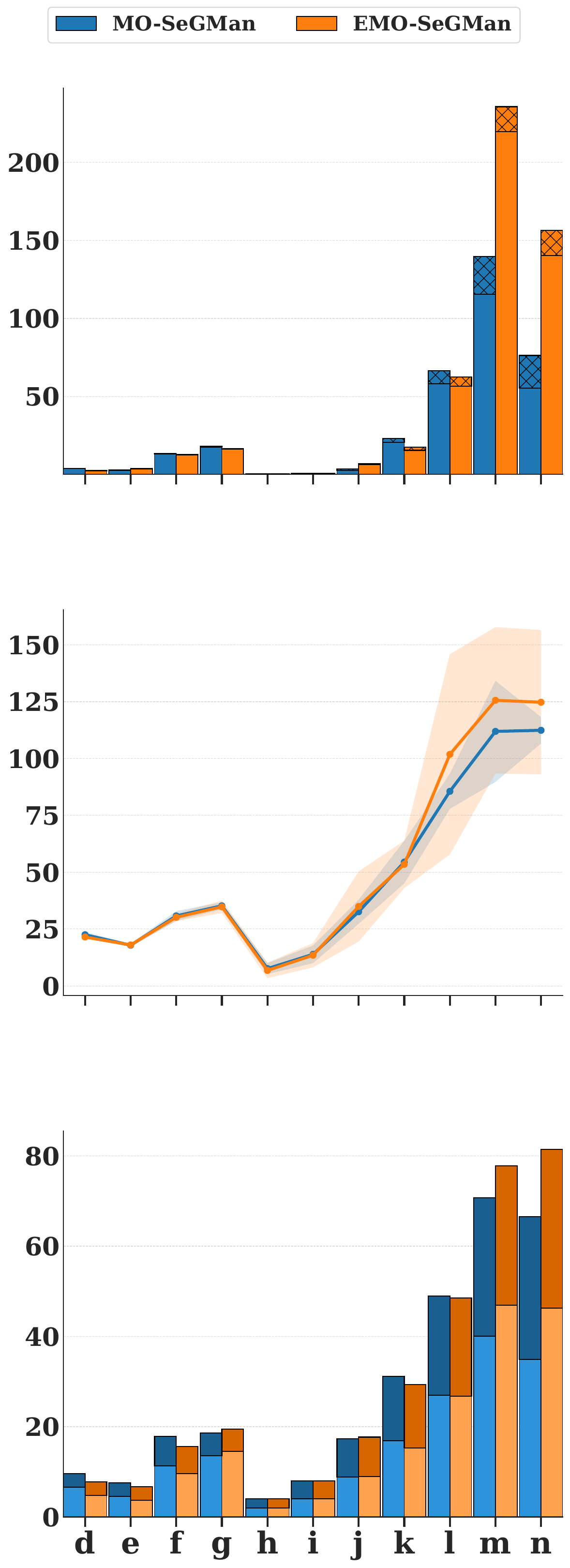}
  \end{minipage}

\caption{Experimental results across the five baselines (SeGMan~\cite{tuncer2025segmansequentialguidedmanipulation}, \texttt{RMO-SeGMan}, \texttt{SMO-SeGMan}, \texttt{GMO-SeGMan}, and \texttt{EMO-SeGMan}). The first row shows solution times, with sequence generation time indicated as shaded regions (in seconds; infeasible cases are reported as percentages above the bars). The second row shows robot travel distance (in meters). The third row shows stacked pick-and-place (PnP) (light) and replanning (dark) counts. Test cases include: (a) O-Room, (b) Slot, (c) Four-Blocks, (d) MO-3-Block, (e) MO-Slot, (f) MO-Puzzle, (g) MO-Order, (h--m) MO-M-Block, and (n) MO-ICRA.}
\label{fig:results}
\end{figure*}

\section{Experiments}
MO-SeGMan is evaluated on 14 complex and constrained rearrangement tasks of varying difficulty (Fig.~\ref{fig:task}), ranging from a single-goal object scenarios (i.e. O-Room) to highly cluttered environments with up to 26 objects (MO-ICRA). In each task, a 2-DoF robot (yellow) manipulates the designated goal objects into their target positions (object silhouettes) while interacting with movable obstacles (white) and other goal objects when necessary. These tasks highlight two key challenges: resolving precedence relationships among objects in non-monotone cases and performing object manipulation in narrow passages among movable objects.

We compare MO-SeGMan against several baselines to evaluate the contributions of its critical components. 
\texttt{RMO-SeGMan} uses the same framework but with a random object placement sequence. 
\texttt{GMO-SeGMan} constructs $G'$ using greedy cycle elimination. 
\texttt{EMO-SeGMan} optimizes the sequence using only Euclidean travel distance as the cost metric. 
\texttt{SMO-SeGMan} does not update $\mathcal{S}$ at intermediate steps. 
Finally, we compare against SeGMan~\cite{tuncer2025segmansequentialguidedmanipulation}, with the object placement sequence provided by MO-SeGMan. 
These baselines allow us to evaluate the impact of each key component on the performance of the proposed framework across the designed cases, as reflected in the evaluation metrics.

The evaluation metrics include success rate, total solution time (including sequence generation), and solution quality. Solution quality is assessed using three key measures: the number of pick-and-place actions (PnP), the replanning count, and the robot’s travel distance, with lower values indicating higher quality. Since MO-SeGMan shares the same motion planning module with \texttt{RMO-SeGMan}, \texttt{GMO-SeGMan}, \texttt{EMO-SeGMan}, and \texttt{SMO-SeGMan}, results for single-goal object cases are omitted for these baselines. Each seeded run is repeated 10 times for statistical significance. For MO-ICRA and MO-M-Block, object and goal locations are randomized across trials, with MO-M-Block evaluated for $M \in {2, 4, 8, 12, 16, 20}$. Any run exceeding 1000 seconds is terminated and recorded as a failure.

\section{Results}
Across all tasks, MO-SeGMan generated feasible motion plans in less time and with higher solution quality compared to the baselines (Fig.~\ref{fig:results}). 
Notably, the motion planner succeeded in producing feasible solutions across all MO-SeGMan variants, even with random $\mathcal{S}$, underscoring the robustness of the overall framework. 
The results confirm that the proposed object placement sequence generation algorithm effectively captures object dependencies while minimizing robot travel distance. Moreover, it is demonstrated that repeated sequence generation is a viable method as the solution time is dominated by the manipulation planning. 
The SGFS enhances efficiency by rapidly identifying feasible relocation points for critical obstacles, reducing overall planning time. 
Finally, the refinement step in adaptive subgoal generation eliminates unnecessary intermediate pick-and-place actions, generating higher-quality solutions.  

For the single goal object cases, both MO-SeGMan and SeGMan generated feasible motion plans, but MO-SeGMan achieved them in less time and with higher solution quality. 
In the O-Room task, the solution times are comparable; however, MO-SeGMan yields significantly fewer PnP actions (\%50 less) due to the proposed filtering step for the adaptive subgoal selection. 
For the multi-goal object cases, MO-SeGMan shows clear advantages, achieving faster solution times and better scalability as the number of objects increases. 
In particularly challenging instances, SeGMan required more than 1000 seconds to generate a motion plan, whereas MO-SeGMan successfully produced solutions well below the time limit (around 120s in hard cases). 
These improvements are primarily attributed to the SGFS method, which efficiently directs the search toward critical obstacles and their feasible relocation points.

Among the MO-SeGMan variants, \texttt{RMO-SeGMan} is the worst-performing in both solution time and quality, which is expected. 
In highly cluttered cases, randomizing the placement sequence often results in premature placements, forcing the planner to regenerate multiple motion plans for the same object, thereby increasing runtime and reducing solution quality.  

\texttt{SMO-SeGMan} and \texttt{GMO-SeGMan} both perform better than the random variant; however, in highly non-monotone cases such as MO-M-Block ($M > 16$) and MO-ICRA, their performance degrades considerably (about \%20 worse). 
For \texttt{SMO-SeGMan}, the main drawback is that changes in the configuration, especially when goal objects are temporarily relocated, are not reflected in $\mathcal{S}$, resulting in a suboptimal placement order. 
In the case of \texttt{GMO-SeGMan}, greedy cycle elimination discards critical dependency edges, leading to increased replanning overhead and reduced solution quality. 
These results justify the time invested in object placement sequence generation in MO-SeGMan, as it substantially reduces total replanning time and improves overall solution quality.  

\texttt{EMO-SeGMan} achieves performance close to MO-SeGMan and even produces solutions slightly faster in smaller instances where $m < 8$. 
However, the resulting motion plans are of lower quality, primarily due to increased robot travel distance in more cluttered cases. 
For small-scale problems, relying solely on Euclidean distances instead of refining $\mathcal{S}$ with RRT-based distances can be a reasonable alternative. 
In highly constrained environments, however, the Euclidean heuristic underestimates the true travel cost, often also increasing solution generation time. 
This again justifies the use of lazy evaluation with iterative RRT-based refinement in MO-SeGMan.

\section{Conclusion}
In this paper, we introduced MO-SeGMan, a multi-objective sequential and guided manipulation planner for rearrangement problems in cluttered and highly constrained environments. 
The planner generates an object placement sequence by minimizing replanning per object and overall robot travel distance, thereby reducing solution time and improving solution quality. The proposed Selective Guided Forward Search (SGFS) selectively relocates only critical obstacles, efficiently exploring configurations for feasible relocations.
In addition, the refinement step in adaptive subgoal generation reduces unnecessary pick-and-place actions, further improving the solution quality. Extensive evaluations across diverse rearrangement tasks show that MO-SeGMan outperforms baseline methods, demonstrating its applicability and scalability to a broad range of settings, such as warehouse and household automation. For future work, we plan to extend MO-SeGMan to multi-robot settings, focusing on task allocation, coordination, and collaboration strategies. 

\bibliographystyle{ieeetr}
\bibliography{main}
\vspace{12pt}

\end{document}